\documentclass[runningheads]{llncs}

\usepackage{url}            
\usepackage{booktabs}       
\usepackage{amsfonts}       
\usepackage{nicefrac}       
\usepackage{microtype}      
\usepackage{xcolor}         
\usepackage{tabularx}
\usepackage{graphicx}
\usepackage{amsmath}
\usepackage{amssymb}
\usepackage{mathtools}
\usepackage{subcaption}
\usepackage{tabulary,multirow,overpic,xcolor}
\definecolor{mygray}{gray}{0.92}
\definecolor{baselinecolor}{gray}{.9}
\newcommand{\baseline}[1]{\cellcolor{baselinecolor}{#1}}
\usepackage{listings}
\usepackage{colortbl}
\usepackage{braket}
\usepackage{array}

\usepackage{float}
\usepackage{booktabs}

\newcolumntype{x}[1]{>{\centering\arraybackslash}p{#1pt}}
\newcolumntype{y}[1]{>{\raggedright\arraybackslash}p{#1pt}}
\newcolumntype{z}[1]{>{\raggedleft\arraybackslash}p{#1pt}}

\newcommand{\tablestyle}[2]{\setlength{\tabcolsep}{#1}\renewcommand{\arraystretch}{#2}\centering\footnotesize}

\begin{document}

\title{SurgMAE: Masked Autoencoders for Long Surgical Video Analysis}

\author{Muhammad Abdullah Jamal and Omid Mohareri}
\institute{Intuitive Surgical Inc., Sunnyvale, CA} 
\maketitle

\begin{abstract}
There has been a growing interest in using deep learning models for processing long surgical videos, in order to automatically detect clinical/operational activities and extract metrics that can enable workflow efficiency tools and applications. However, training such models require vast amounts of labeled data which is costly and not scalable. Recently, self-supervised learning has been explored in computer vision community to reduce the burden of the annotation cost. Masked autoencoders (MAE) got the attention in self-supervised paradigm for Vision Transformers (ViTs) by predicting the randomly masked regions given the visible patches of an image or a video clip, and have shown superior performance on benchmark datasets. However, the application of MAE in surgical data remains unexplored. In this paper, we first investigate whether MAE can learn transferrable representations in surgical video domain. We propose SurgMAE, which is a novel architecture with a masking strategy based on sampling high spatio-temporal tokens for MAE. We provide an empirical study of SurgMAE on two large scale long surgical video datasets, and find that our method outperforms several baselines in low data regime. We conduct extensive ablation studies to show the efficacy of our approach and also demonstrate it's superior performance on UCF-101 to prove it's generalizability in non-surgical datasets as well.


\end{abstract}
\section{Introduction}\label{sec:intro}
Robotic-assisted surgery (RAS) has been widely adopted for many surgical procedures since it allows surgeons to perform operations with more precision and provides benefits such as fast post-operative recoveries, less blood loss and shorter hospitalization~\cite{Sheetz2020}. However, the adaptation of RAS is still not ubiquitous due to barriers such as cost, training, and OR workflow complexities~\cite{Catchpole2015SafetyEA}.
Many component technologies have been recently introduced to address such issues. Methods such as automatic activity recognition in Operating room (OR)~\cite{aidean,Schmidt}, scene understanding and context awareness in the OR~\cite{Li}, and endoscopic video workflow recognition~\cite{zhang2021real} have shown the potential of enabling digital tools that can improve workflow processes for the surgeon and OR staff. 

The focus of this paper is automatic surgical activity recognition (SAR) which is the task of detecting activities or phases temporally in long videos.~\cite{aidean,OR-AR} introduced a new dataset called OR-AR which consists of long videos collected from different ORs and employ two-stage training. However, these approaches are fully supervised and require clinical data that is manually annotated by medical experts which can impede the scalability of these models. Indeed, we want to have models that can understand the surgical workflow in a scalable fashion either at surgery level or OR level. This requires us to have machine learning models that are data-efficient in nature.

In this paper, we leverage the unlabeled long surgical videos and propose a new self-supervised learning (SSL) approach based on masked autoencoders. SSL learns generic representations on unlabeled data that are transferable to various downstream tasks. There are two different approaches used in SSL. In contrastive learning, model takes different augmentations or views of an image and then use contrastive loss to pull them together in the embedding space while pushing away the embeddings of different images. On the other hand, masked autoencoders (MAE)~\cite{MAE,maskfeat,videomae} take an input (either image or video), patchify it and pass it into patch embedding to generate a set of tokens. A high percentage of tokens are generally dropped and the remaining tokens are passed to the Vision Transformers (ViT)~\cite{Vit}. Then, the tokens embeddings concatenated with learnable masked tokens are passed to the decoders to reconstruct the masked patches. MAEs have recently gained more momentum in SSL paradigm because of less inductive bias, high masking ratio and have consistently been outperforming contrastive learning based approaches. To the best of our knowledge, little or no work has been done in the surgical domain using MAE. We first investigate whether MAE can learn useful representations during pre-training for surgical video datasets. We train MAEs using different masking strategies such as random~\cite{MAE}, tube~\cite{videomae} and frame~\cite{videomae} masking. We empirically found that the random masking strategy works best. However, all the tokens are not informative as there are a lot of redundant information in the video and random masking strategy can select tokens from uninformative regions. To tackle this challenge, we propose a new masking strategy that samples tokens from high spatio-temporal regions. Our masking strategy selects tokens based on the distance in the embedding space while discarding tokens from low information regions (background, redundant frames etc.). We empirically show the efficacy of our new masking strategy on multiple datasets to prove out that it selects tokens from high informative regions, resulting in learning more useful representations for downstream tasks. In summary, the main contributions of our paper are:
\begin{itemize}
    \item To the best of our knowledge, this is the first work to explore MAE based pre-training using Vision transformers for long surgical video datasets.
    \item We propose a new and effective sampling strategy to select tokens from high informative regions while discarding tokens from uninformative spatio-temporal regions in the video clip.
    \item We conduct in-depth ablation studies, and empirically demonstrate that our approach outperforms several masking strategies on two large-scale surgical video datasets.
    \item We show that our approach is generalizable by demonstrating it's efficacy on two surgical datasets as well as UCF-101~\cite{ucf101}, a benchmark video dataset in computer vision community.
\end{itemize}

\section{Related Work}
Annotating massive amounts of surgical video data temporally and spatially requires manual work from medical experts and is impractical and expensive. To put our work in the context of this application and prior works, we briefly review the surgical video understanding, OR workflow analysis, and self-supervised learning paradigms. Please see supplementary materials for extended literature review.
\paragraph{\textbf{Surgical Video Understanding.}} Datasets such as Cholec80~\cite{Cholec80} and Cataract-101~\cite{cataracts} have allowed us to make advancement in surgical phase recognition mainly in laparoscopic and ophthalmological videos. Recent approaches~\cite{aidean,techno} are mainly supervised and consist of two stages. 
\paragraph{\textbf{Self-supervised Learning (SSL).}} Recently, contrastive learning~\cite{cvrl,Mocov3,coclr,videomoco} approaches and masked autoencoder~\cite{videomae,MAE,maskfeat,imae,mcmae} are mainly used in SSL to learn better visual representations from large scale unlabeled datasets.
\section{Method}\label{sec:method}
In this section, we will first review Masked Autoencoders for images, and then briefly discuss the current trends of MAEs for video representation learning. Finally, we will discuss SurgMAE architecture which consists of our token sampling strategy.
\subsection{Masked Autoencoders for images}
MAE for images~\cite{imae} aims to reconstruct the masked patches in images. The image is passed to a 2D convolution layer to extract patch embeddings. Then, it randomly masks these patch embeddings (tokens) with high ratio while the remaining ones are input into the ViT encoder. Finally, the tokens from the encoder concatenated with the learnable mask tokens are fed into the decoder to reconstruct the masked patches. The L2 loss (mean squared error) is used as a training loss between the prediction and the normalized masked tokens in the pixel space.
\subsection{Masked Autoencoders for videos}
VideoMAE~\cite{videomae} and MAE~\cite{MAE} extend the work of ImageMAE~\cite{imae} to the video domain. VideoMAE~\cite{videomae} proposed a ``tube'' masking strategy which randomly masks in 2D spatial dimension and then extend it to the temporal dimension to create a tube. MAE~\cite{MAE} proposed ``random'' masking which randomly masks tokens in space and time. They also explore ``frame'' masking strategy which masks future or past frames, and consequently makes the task more challenging to learn meaningful representations for downstream tasks. Given the redundant nature of video data, these approaches use high masking ratio (75\% to 90\%) during pre-training. Although, MAE~\cite{MAE} observed that the random masking performs the best among various masking strategies, it can also sample tokens from background or uninformative regions which might not be useful for learning useful representation. Given these challenges, we propose a new sampling strategy to sample tokens from high informative regions.
\begin{figure*}
    \centering
    \includegraphics[width=0.70\textwidth]{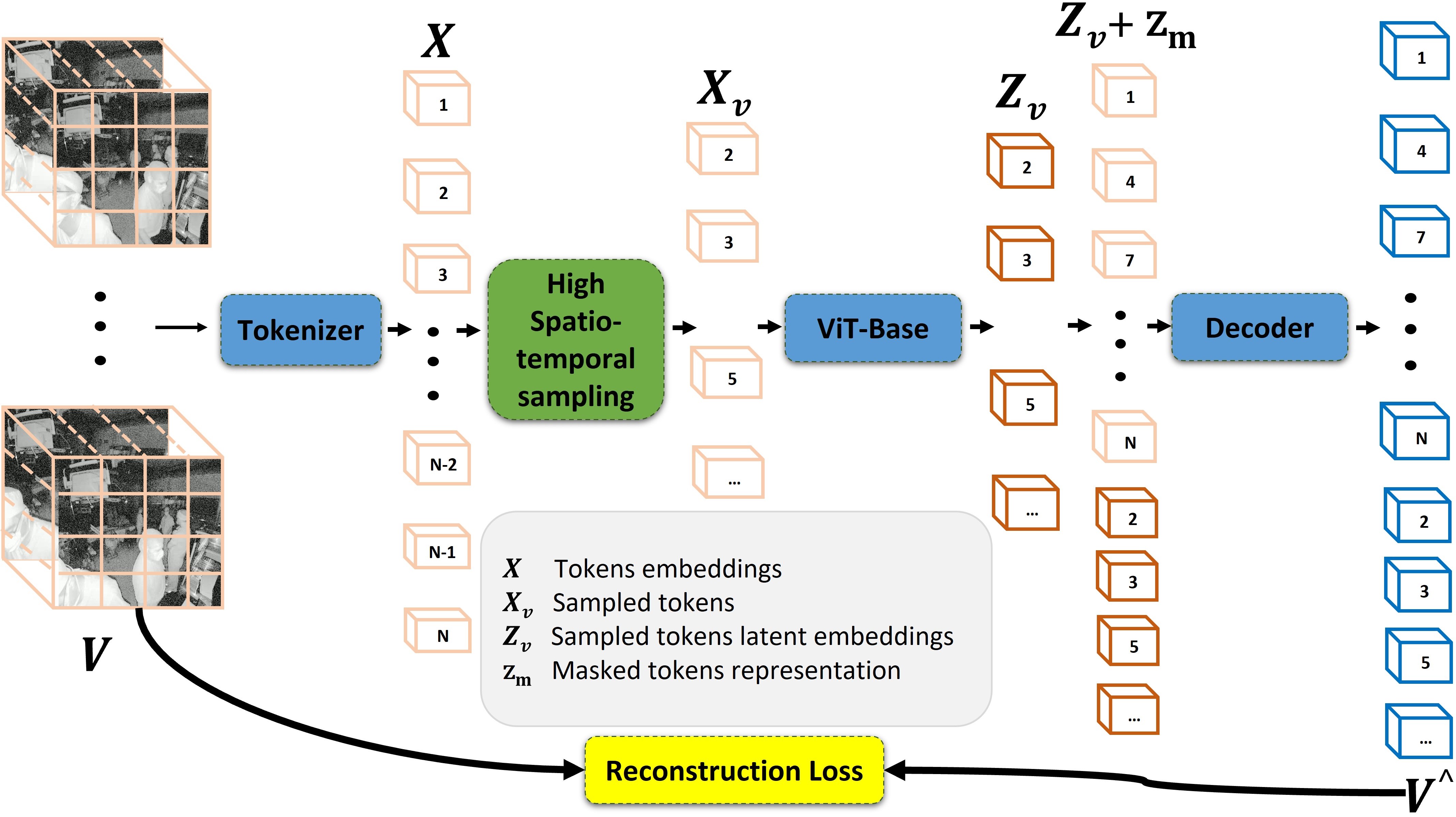}
    \caption{SurgMAE samples high spatio-temporal tokens based on the distance between adjacent frames and reconstructs the masked ones using ViT encoder-decoder.}
    \label{fig:approach}
    \vspace{-3.0em}
\end{figure*}
\subsection{SurgMAE}
Figure~\ref{fig:approach} shows the architecture of SurgMAE. Our goal is to learn video representations under masked autoencoder paradigm using high spatio-temporal tokens. Given a video clip $\textbf{V}$ of size $T$ x $3$ x $H$ x $W$ where $T$ is the number of frames in the clip, $H$ and $W$ are height and width of the frame and $3$ corresponds to the number of channels in the frame, we first pass it to a 3D convolutional layer with a patch size of $2$x$16$x$16$ to extract $N =$  $\frac{T}{2}$x$\frac{H}{16}$x$\frac{W}{16}$ tokens of dimension \textbf{\textit{d}}. Let $\mathbf{X}_{i}$ be the token embedding of two adjacent frames ($i, i+1$) of the video clip. We use the euclidean norm between the token embedding $\mathbf{X}_{i+1,k}$ and the token embedding of the previous frame $\mathbf{X}_{i,k}$ over the same 2D position to compute if the token $\mathbf{X}_{i+1,k}$ belongs to high information region or not. Then, we only sample tokens with high distance value based on the masking ratio. The intuition is if the spatial location of object changes in the next frame, then it should be accounted as high information token.
Next, we follow MAE~\cite{MAE} to adapt separable positional embedding, one for space, and other for the time. For non-surgical data, we follow~\cite{videomae} for adding positional embeddings into the tokens. After positional embeddings are added, the sampled tokens $\mathbf{X}_{v}$ are passed to the encoder to extract latent representations $\mathbf{Z}_{v}$. The latent embeddings are then concatenated with learnable masked tokens $\mathbf{z}_{m}$. Finally, the positional embeddings are added, and passed to the decoder to reconstruct the masked patches $\hat{\textbf{V}}$. Following~\cite{videomae,MAE}, we use mean squared error (MSE) as a loss function between the prediction and the normalized RGB pixel values:
\begin{equation}
\mathcal{L} = \frac{1}{\omega} \sum_{p \in \omega} \vert \vert  \textbf{V} (p) - \hat{\textbf{V}}(p) \vert \vert_{2}
\label{eq:loss}
\end{equation}
where $p$ is the token index, $\omega$ is the set of masked tokens.  Please see section~\ref{subsec:ablation} for more ablation study on the effect of different loss functions.
We remark that our masking strategy would allow the model to sample tokens based on the spatio-temporal information from each video clip as compared to random masking and have empirically shown the superior performance of the downstream tasks in the section~\ref{subsec:results}.


\section{Experiments}\label{sec:exps}
\paragraph{\textbf{Datasets.}} We evaluate SurgMAE on three datasets: OR-AR~\cite{OR-AR} and OR-ARv2, Cataract-101~\cite{cataracts}, and UCF-101~\cite{ucf101}. OR-AR is a surgical dataset consisting 820 full videos captured using IR cameras placed in two ORs in a single hospital. OR-ARv2 is an extended version of OR-AR dataset that has videos collected from three new hospitals. This dataset includes 1302 OR videos and has been collected under institutional research board (IRB) approvals. Please see the supplementary materials for more details on the datasets.
\paragraph{\textbf{Implementation details.}} We use Vision transformer-Base (ViT-B) with joint-space time attention as the backbone following prior work~\cite{videomae}. We set the input number of frames to 16 and sampling rate of 4.0, and set the patch size of 2x16x16 which generates 1568 tokens for an input video clip with a size of 224x224. Please see the supplementary materials for more implementation details.
\paragraph{\textbf{Evaluation}.} In order to evaluate the pre-trained models for surgical datasets, we fine-tune ViT-B model on video clips similar to~\cite{aidean}, and then extract features for full videos from ViT-B to train Bidirectional Gated Recurrent Unit (Bi-GRU) to detect surgical activities in surgical video datasets. For UCF-101, we only perform end-to-end fine-tuning of the backbone. We use mean average precision (mAP) and top-1 accuracy as evaluation metrics.



\begin{table}[t]
\centering
\caption{Ablation studies on \textbf{OR-AR~\cite{OR-AR}} under low-data (5\% labeled data) regime setting. We use ViT-B as a backbone for all the experiments.}
\subfloat[
\textbf{Different Masking Ratio}. SurgMAE works well with high masking ratio. Models are trained for 400 epochs
\label{tab:mratio}
]{
\centering
\begin{minipage}{0.27\linewidth}{\begin{center}
\tablestyle{4pt}{1.05}
\begin{tabular}{x{24}x{22}}
ratio & mAP \\ 
\hline

\hline

\hline
0.95  &  63.37   \\ \hline
0.90  &  \baseline{\textbf{64.97}}   \\ \hline
0.85  &  62.26   \\ \hline
0.80  &  61.01  
\end{tabular}
\end{center}}\end{minipage}
}
\hspace{2em}
\subfloat[
\textbf{Decoder Depth}. SurgMAE performs the best with 4 blocks of decoder. Models are trained for 800 epochs with masking ratio of 90\%.
\label{tab:decoder}
]{
\begin{minipage}{0.27\linewidth}{\begin{center}
\tablestyle{4pt}{1.05}
\begin{tabular}{x{22}x{24}}
blocks & mAP \\ 
\hline

\hline

\hline
1  &   61.35   \\ \hline
2  &   65.84  \\ \hline
4  &   \baseline{\textbf{67.86}} \\ \hline
8  &   62.75 
\end{tabular}
\end{center}}\end{minipage}
}
\hspace{2em}
\subfloat[
\textbf{Mask sampling}. SurgMAE outperforms random, frame and tube masking.
\label{tab:masking}
]{
\begin{minipage}{0.24\linewidth}{\begin{center}
\tablestyle{4pt}{1.05}
\begin{tabular}{y{50}x{20}}
case & mAP \\
\hline

\hline

\hline
random  &  66.58   \\ \hline
tube  &  65.57   \\ \hline
frame  &  63.44   \\ \hline
SurgMAE  & \baseline{\textbf{68.91}}
\end{tabular}

\end{center}}
\end{minipage}
}
\\
\centering
\subfloat[
\textbf{Pre-training epochs}. Better performance achieves during fine-tuning when pre-trains for more epochs.
\label{tab:abl_epochs}
]{
\centering
\begin{minipage}{0.43\linewidth}{\begin{center}
\tablestyle{4pt}{1.05}
\begin{tabular}{x{25}x{20}}
epochs & mAP \\
\hline

\hline

\hline
400  &  64.97   \\ \hline
600  &  65.89   \\ \hline
800  &   67.86  \\ \hline
1600  &  \baseline{\textbf{68.91}}
\end{tabular}
\end{center}}\end{minipage}
}
\hspace{1.5em}
\subfloat[
\textbf{Loss function}. SurgMAE performs best with MSE loss and normalization. Models are trained for 800 epochs with masking ratio of 90\%.
\label{tab:abl_loss}
]{
\begin{minipage}{0.45\linewidth}{\begin{center}
\tablestyle{4pt}{1.05}
\begin{tabular}{y{90}x{24}}
case & mAP \\
\hline

\hline

\hline
MSE (w / norm )  & \baseline{\textbf{67.86}}    \\ \hline
MSE (wout / norm ) & 63.41    \\ \hline
L1 (w / norm ) & 64.06    \\ \hline
L1 (wout / norm ) &  62.14  
\end{tabular}
\end{center}}\end{minipage}
}
\label{tab:ablations} 
\end{table}
\subsection{Ablation Study}\label{subsec:ablation}
In this subsection, we perform in-depth ablation studies of SurgMAE on OR-AR~\cite{OR-AR} dataset. We pre-trained ViT-B on the OR-ARv2 dataset and then fine-tune it under low-data regime setting (5\% labeled data) for evaluation. We report these studies in Table~\ref{tab:ablations}.
\paragraph{\textbf{Masking ratio.}} Table~\ref{tab:mratio} shows the performance of SurgMAE on different masking ratios. It shows that SurgMAE performs well with a high masking ratio which makes the pre-training fast and less memory intensive. Surprisingly, masking ratio of 95\% achieves 63.37\% mAP which is in line with the fact that SurgMAE samples more high spatio-temporal tokens and requires fewer tokens during pre-training to achieve a reasonable performance. We observe a drop in fine-tuning performance when pre-training the model using lower masking ratio. We hypothesize that lower masking ratios sample more redundant patches which results in poor generalization.
\paragraph{\textbf{Decoder design.}} Table~\ref{tab:decoder} shows the performance of SurgMAE on different blocks of the decoder. We observe that the performance increases when increasing the depth of the decoder from 1 block to 4 blocks. However, with much deeper decoder, we see a performance degradation which is in accordance with the observation made in recent masked autoencoder approaches~\cite{MAE,videomae}.
\paragraph{\textbf{Masking strategy.}} Table~\ref{tab:masking} compares the performance of SurgMAE with the recent masking strategies. We observe that the random masking outperforms tube masking with high masking ratio. However, frame masking which masks out future or past frames performs poorly compared to tube or random masking. Same observation has been made in~\cite{MAE}. \textbf{SurgMAE} which samples high spatio-temporal tokens yields the best performance with high masking ratio (68.91\% with 90\% ratio).
\paragraph{\textbf{Pre-training epochs.}} Next, we show the impact of pre-training epochs on the fine-tuning performance in Table~\ref{tab:abl_epochs}. We observe an increase in the performance (64.97\% mAP to 67.86\% mAP) when number of epochs goes from 400 to 800. If we further pre-train the model with 1600 epochs, we achieve our best performance, but it comes with a cost of more pre-training time. 
\paragraph{\textbf{Reconstruction target.}} Table~\ref{tab:abl_loss} compares the performance of SurgMAE using different loss function. We observe that MSE loss performs better compared to L1 loss. We also observe that per-patch normalized pixels yields better results compared to using raw pixel value which is on par with observation made in recent masked autoencoder approaches~\cite{MAE,videomae}.  
\begin{table}[h!]
\centering
\vspace{-1em}
\caption {\label{tab:OR-AR} \textbf{Comparison of SurgMAE with the other state-of the art methods under different data-regime setting on the OR-AR dataset~\cite{OR-AR}}. We pre-train ViT-B using SurgMAE for 1600 epochs with high masking ratio of 90\%.}
\resizebox{0.85\textwidth}{!}{%
\begin{tabular}{c|c|c|c|c|c|c|c}
\hline

\hline

\hline
Methods  & Masking  & Backbone & Pre-train    & 5\%  & 10\%  & 20\% & 100\% \\ 
\hline

\hline

\hline
MaskFeat~\cite{maskfeat}      & Random                                                                  & MViT-S    & Kinetics-400 &62.35  & 78.88 &-  &- \\ \hline
MAE~\cite{MAE}       & Random                                                                  & ViT-B    & Kinetics-400 & 64.66 & 81.48 &84.93 & 94.76\\ \hline
MAE~\cite{MAE}       & Random                                                                  & ViT-B    & OR-ARv2 & 66.58
 & 81.87 & 84.97& \textbf{96.30}\\ \hline
MAE~\cite{MAE}       & Frame                                                                  &  ViT-B        & OR-ARv2    & 63.44 & 78.89 &81.45& -\\ \hline
VideoMAE~\cite{videomae} & Tube                                                                    & ViT-B    & OR-ARv2    & 65.57 & 81.74& 83.89 &94.87\\ \hline
\textbf{SurgMAE}  & \begin{tabular}[c]{@{}c@{}}high spatio-\\temporal sampling\end{tabular} & ViT-B    & OR-ARv2    & \textbf{68.91} &\textbf{82.14} &\textbf{86.29} & 95.60\\
\hline

\hline

\hline

Swin-B+BiGRU~\cite{OR-AR} & -                                                                    & Swin-B   & Kinetics-400    & - & - & - & 95.13\\ \hline

\end{tabular}
}%
\end{table}

\vspace{-2.0em}
\subsection{Main Results}\label{subsec:results}
We compare SurgMAE with the recent masked autoencoder and contrastive learning approaches for video domain. Please see supplementary materials for more details on the baselines and the results on the Cataract-101. Table~\ref{tab:OR-AR} shows the comparison of various approaches under different data regimes for OR-AR~\cite{OR-AR}. We can observe that SurgMAE has a clear advantage over other masking approaches under low-data regime setting. More specifically, SurgMAE achieves 68.91 mAP when fine-tuned using 5\% labeled data which shows that it is a more data-efficient learner than the recent sota approaches. Among other approaches, we observe that MAE~\cite{MAE} with random masking performs reasonably well under low-data regime setting and even performs slightly better than SurgMAE when fine-tune using full labeled dataset (96.30\% vs 95.60\% mAP). Moreover, both random masking and SurgMAE performs better than the fully supervised results obtained using pre-trained Swin transformer-Base~\cite{swin} (Swin-B) as a backbone which shows the effectiveness of masked autoencoders when pre-training on in-domain large scale dataset. 

\begin{table}[h!]
\centering
\vspace{-2em}
\caption {\label{tab:extended_OR-AR} \textbf{Comparison of SurgMAE with the other state-of the art methods on the OR-ARv2 dataset}.}
\resizebox{0.65\textwidth}{!}{%
\begin{tabular}{c|c|c|c|c}
\hline

\hline

\hline
Methods  & Masking  & Backbone & Pre-train    & mAP   \\ 
\hline

\hline

\hline
MAE~\cite{MAE}      & Random                                                                  & ViT-B    & Kinetics-400 & 92.06 \\ \hline
MAE~\cite{MAE}       & Random                                                                  & ViT-B    & OR-ARv2 & 92.70 \\ \hline
MAE~\cite{MAE}       & Frame                                                                  &  ViT-B     & OR-ARv2    & 91.87 \\ \hline
VideoMAE~\cite{videomae} & Tube                                                                    & ViT-B    & OR-ARv2   & 92.36 \\ \hline
\textbf{SurgMAE}  & \begin{tabular}[c]{@{}c@{}}high spatio-\\temporal sampling\end{tabular} & ViT-B    & OR-ARv2    & \textbf{93.11} \\ \hline

\end{tabular}
}%
\vspace{-0.5em}
\end{table}

Table~\ref{tab:extended_OR-AR} compares the performance of SurgMAE on OR-ARv2 on full labeled dataset. We see the same observation that SurgMAE yields the best performance (93.11\% mAP) compared to other masking strategies which empirically verify the effectiveness of sampling high spatio-temporal tokens during pre-training. Moreover, similar to what we observe in OR-AR results, MAE~\cite{MAE} with random masking performs better than the tube and frame masking.
\vspace{-1em}
\begin{table}[h!]
\centering
\vspace{-1em}
\caption {\label{tab:ucf101} \textbf{Comparison of SurgMAE with the other state-of the art methods on UCF-101~\cite{ucf101}}.}
\resizebox{0.70\textwidth}{!}{%
\begin{tabular}{c|c|c|c|c|c}
\hline

\hline

\hline
Methods  & Masking  & Backbone & Pre-train & Frames & Top-1 \\
\hline

\hline

\hline
Scratch  & -                                                                       & ViT-B    & UCF-101   & 16     & 51.4  \\ \hline
MoCov3~\cite{Mocov3}   & -                                                                       & ViT-B    & UCF-101   & 16     & 81.7  \\ \hline
VideoMAE~\cite{videomae} & Tube                                                                    & ViT-B    & UCF-101   & 16     & 91.2 \\ \hline
\textbf{SurgMAE}  & \begin{tabular}[c]{@{}c@{}}high spatio-\\temporal sampling\end{tabular} & ViT-B    & UCF-101   & 16     & \textbf{92.1}  \\ \hline
\end{tabular}
}%
\end{table}

\vspace{-2.1em}
\paragraph{\textbf{UCF-101.}} Finally, to empirically test the generalizability of SurgMAE on non-surgical dataset, we run experiments on UCF-101. We report the top-1 accuracy in Table~\ref{tab:ucf101}. It can be clearly seen that SurgMAE outperforms VideoMAE~\cite{videomae} achieving 92.1\% top-1 accuracy compared to 91.2\% which shows the efficacy of visible tokens from high information regions for masked autoencoder. For UCF-101, we also sample tokens from the static regions on the top of high spatio-temporal tokens and didn't mask all the remaining tokens for the prediction task. We use a high masking ratio of 80\% as compared to 75\% ratio used in VideoMAE~\cite{videomae} making it less memory and computational intensive. 

\section{Conclusion}
In this paper, we investigate masked autoencoder based pre-training techniques for long surgical videos to learn better video representations. We propose SurgMAE, an adapted version of MAE for surgical videos, with a simple and effective token sampling strategy which samples tokens from high spatio-temporal regions to alleviate issues with random masking methods commonly used in current MAE approaches. We empirically show that our approach outperforms other masking strategies with ViT-B model on two surgical and one non-surgical (UCF-101) datasets to prove it's superior representation learning capability.

\bibliographystyle{splncs04}
\bibliography{references}

\newpage
\appendix
\section*{Appendices}
\addcontentsline{toc}{section}{Appendices}
\renewcommand{\thesubsection}{\Alph{subsection}}


\section{Datasets Details} \label{dataset}
Cataract-101~\cite{cataracts} is a small dataset consists of 101 surgical videos and 10 phases. UCF-101~\cite{ucf101} is also a relatively small dataset which consists of 9.5k videos in training and 3.5k in testing.
OR-AR~\cite{OR-AR} and OR-ARv2 consist of 9 phases and contain 61 type of surgeries such as Lobectomy, Colectomy, Umbilical Hernia Repair etc. We split the dataset into 80-20 train-test split. 

\section{Baselines} \label{baselines}
We directly compare SurgMAE to the recent masked autoencoders that includes MaskFeat~\cite{maskfeat}, VideoMAE~\cite{videomae}, MAE~\cite{MAE}. We also compare with SwAV~\cite{swav} and SimCLR~\cite{simclr} for Cataract-101 experiments. Moreover, we add MoCov3~\cite{Mocov3} as one of the baselines for UCF-101 experiments following VideoMAE~\cite{videomae}.


\section{Implementation Details} \label{implementations}
We conduct our experiments on 8 NVIDIA Tesla V100 GPUs. We follow~\cite{videomae,MAE} for pre-training (see Table~\ref{tab:pretrain}) and fine-tuning (see Table~\ref{tab:fine-tune}) settings.

\begin{table}[h]
    \centering
    \tablestyle{2.0pt}{1.04}
    \caption{Pre-training setting on OR-ARv2, UCF-101 and Cataract-101 datasets.}
    \resizebox{0.65\textwidth}{!}{%
    \begin{tabular}{l|ccc}
    \hline
    
    \hline
    
    \hline
    Configuration & OR-ARv2 & UCF-101  & Cataract-101\\
    \hline
    
    \hline
    
    \hline
    Optimizer & \multicolumn{3}{c}{Adamw}\\
    Optimizer betas & \multicolumn{3}{c}{\{0.9, 0.95\}}\\
    Base learning rate & 1e-4 & 1e-3 & 1e-4\\
    Weight decay & \multicolumn{3}{c}{5e-2}\\
    Learning rate schedule & \multicolumn{3}{c}{cosine decay} \\
    gradient clipping & 0.02 & None & 0.02\\
    Warmup epochs & \multicolumn{3}{c}{40}\\
    Epochs & 1600 & 3200 & 800\\
    Flip augmentation & True & True & False\\
    Augmentation & \multicolumn{3}{c}{MultiScaleCrop}\\
    Num of Frames & \multicolumn{3}{c}{16}\\
    sampling rate & \multicolumn{3}{c}{4.0}\\
    \hline
    
    \hline
    
    \hline
    \end{tabular}
    }%
    \label{tab:pretrain}
\end{table}

\begin{table}[t]
    \centering
    \tablestyle{2.0pt}{1.04}
    \caption{Fine-tune setting on OR-AR, UCF-101 and Cataract-101 datasets.}
    \resizebox{0.65\textwidth}{!}{%
    \begin{tabular}{l|ccc}
    \hline
    
    \hline
    
    \hline
    Configuration & OR-AR & UCF-101  & Cataract-101\\
    \hline
    
    \hline
    
    \hline
    Optimizer & \multicolumn{3}{c}{Adamw}\\
    Optimizer betas & \multicolumn{3}{c}{\{0.9, 0.95\}}\\
    Base learning rate & 6e-4 & 1e-3 & 6e-4\\
    Weight decay & \multicolumn{3}{c}{5e-2}\\
    Learning rate schedule & \multicolumn{3}{c}{cosine decay} \\
    Warmup epochs & \multicolumn{3}{c}{5}\\
    Epochs & \multicolumn{3}{c}{100}\\
    Flip augmentation & True & True & False\\
    Mixup & None & 0.8 & None\\
    CutMix & None & 1.0 & None\\
    drop path & 0.1 & 0.2 & 0.1\\
    drop out & 0.0 & 0.5& 0.0 \\
    Layer-wise lr decay & 0.65 & 0.70 & 0.65\\
    \hline
    
    \hline
    
    \hline
    Temporal Model learning rate & 1e-3 & None & 1e-3\\
    Temporal Model Epochs & 25 & None & 25\\
    \hline
    
    \hline
    
    \hline
    \end{tabular}
    }%
    \label{tab:fine-tune}
\end{table}

\begin{table}[t]
\centering
\resizebox{0.85\textwidth}{!}{%
\begin{tabular}{c|c|c|c|c}
\hline

\hline

\hline
Methods  & Masking  & Backbone & Pre-train    & mAP   \\ 
\hline

\hline

\hline
SwAV~\cite{swav}     & -        & ViT-B    & Cataracts    & 83.61 \\ \hline
SimCLR~\cite{simclr}   & -                                                                       & ViT-B    & Cataracts    & 83.20 \\ \hline
MAE~\cite{MAE}      & Random                                                                  & ViT-B    & Cataracts    & 86.43 \\ \hline
VideoMAE~\cite{videomae} & Tube                                                                    & ViT-B    & Cataracts    & 85.05 \\ \hline 
\textbf{SurgMAE}  & \begin{tabular}[c]{@{}c@{}}high spatio-\\temporal sampling\end{tabular} & ViT-B    & Cataracts    & \textbf{87.78} \\ 
\hline

\hline

\hline
MAE~\cite{MAE}      & Random                                                                  & ViT-B    & Kinetics-400 & \textbf{92.85} \\ \hline
\end{tabular}
}%
\caption {\label{tab:cataracts} \textbf{Comparison of SurgMAE with the other state-of the art methods on Cataract-101~\cite{cataracts}}. We pre-train ViT-B using SurgMAE for 800 epochs with high masking ratio of 90\%. For evaluation, we report mean average precision of Bi-GRU during fine-tuning stage.}
\end{table}

\section{Results on Cataract-101} \label{cataract}
We report the mAP of Bi-GRU for various approaches in Table~\ref{tab:cataracts}. We carefully follow the training practices to pre-train ViT-B for SimCLR and SwAV to avoid collapse issue. We observe that Cataract-101, being a relatively small dataset, is more challenging to pre-train vision transformers which is on par with the observation found in VideoMAE~\cite{videomae}. Nevertheless, SurgMAE still outperforms other masking approaches which makes it more data-efficient approach for self-supervised pre-training. We also find out that by fine-tuning from a ViT-B pre-trained on large-scale dataset (kinetics-400~\cite{kinetics}) yields the best performance which is generally a standard practice for such small datasets.

\section{Extended Related Work}
In this section, we first review the recent trends in the surgical video understanding and Operating Room (OR) workflow analysis, and then we will briefly discuss the trends and approaches in self-supervised learning paradigm.
\paragraph{\textbf{Surgical Video Understanding.}}~\cite{techno} proposed to use temporal convolution neural network (TCN)~\cite{TCN} on the top of frame-wise features extracted from ResNet-18~\cite{ResNet}. Surgical activity recognition is not just limited to endoscopic videos, but it has been studied for operating room (OR) workflow analysis. The first large scale dataset (OR-AR) was first introduced in ~\cite{aidean} which also proposed a supervised model consists of I3D and Bi-GRU as backbone and temporal model respectively. The dataset was later extended in~\cite{OR-AR}.~\cite{Schmidt} leverages the multiple views of OR-AR dataset, and proposed a new attention module to smartly fuse those views. There are other data-driven based approaches~\cite{Kadkhodamohammadi17} for OR workflow analysis which uses multi-view RGBD dataset~\cite{mvor} for clinician detection and human pose estimation. However, there is little or no work has been done in data-efficient surgical activity recognition. Recently~\cite{Jamal} proposes an unsupervised approach based on clustering which fuses multi-modal data collected from OR. 
\paragraph{\textbf{Self-supervised Learning.}} Self-supervised learning based approaches for learning good video representations have been studied in the literature. Recently, contrastive learning based approaches~\cite{cvrl,Mocov3,coclr,videomoco} has been proposed to learn better visual representation. These approaches generally require larger batch sizes, extra memory component and data augmentations.
\paragraph{\textbf{Masked visual modeling.}} Masked visual modeling leverages the idea from masked language modeling used in bidirectional encoder (BERT)~\cite{BERT} and Generative Pre-Training (GPT)~\cite{GPT}. iGPT~\cite{iGPT} follows GPT to process the pixel in sequential manner which shows that the masked pixel prediction can be performed. Recently, Vision Transformers (ViT)~\cite{Vit} are designed which convert the patches into tokens to learn visual representations. Following the success of ViTs, several masked autoencoder based self-supervised approaches~\cite{MAE,maskfeat,imae,mcmae} has been proposed. Masked Image Modeling~\cite{imae} is a big success and an alternate approach to contrastive learning to learn useful image representations. Similarly, several masked autoencoder based approaches~\cite{videomae,MAE,adamae} are designed for video domain to learn spatio-temporal representations. MAE~\cite{MAE} uses asymmetric encoder-decoder ViT with random masking during pre-training. VideoMAE~\cite{videomae} proposes tube masking strategy while AdaMAE~\cite{adamae} proposes a sampling network to sample tokens from high spatio-temporal regions which is trained end-to-end with ViT encoder using reinforcement learning. MaskFeat~\cite{maskfeat} instead of predicting masked patches, predicts the features of the masked tokens. These approaches have shown to use high masking ratio (75\% to 95\%) during pretraining as opposed to 60\% masking ratio in image domain.

 \label{related}

\end{document}